\documentclass{article}
\usepackage[utf8]{inputenc}
\usepackage{amsmath}
\usepackage{commath}
\usepackage{multicol}
\usepackage{paracol}
\usepackage{cite}
\usepackage{apacite}
\usepackage[margin=0.5in]{geometry}
\usepackage{graphicx}
\usepackage{float}
\usepackage{amsfonts}
\usepackage{amssymb}
\usepackage{abstract}

\renewcommand{\today}{\ifcase \month \or January\or February\or March\or %
April\or May \or June\or July\or August\or September\or October\or November\or %
December\fi, \number \year}

\title{Comparison of PCA with ICA from data distribution perspective}
\author{Miron Ivanov }

\begin{document}
\renewcommand{\abstractname}{}
\renewcommand{\absnamepos}{empty}

\maketitle

\begin{abstract}
We performed an empirical comparison of ICA and PCA algorithms by applying them on  two simulated noisy time series with varying distribution parameters and level of noise. In general, ICA shows better results than PCA because it takes into account higher moments of data distribution. On the other hand, PCA remains quite sensitive to the level of correlations among signals.  
\end{abstract}

\section{Introduction}

\indent ICA algorithm has a similar procedure to PCA in terms of minimizing the objective function and deriving optimal weights. We used iterative version of PCA and ICA exactly to be able to notice that fact. 
\\
\indent PCA has a property of ranking the basis vectors. The one that has the highest eigenvalue will correspond to the principal component which has the highest variance and which explains the largest part of the data and etc. ICA does not have such property. In fact, the algorithm finds a basis up to a scaling factor which means that a basis vector might be randomly multiplied by -1. ICA implies some assumptions on the distribution of data: all sources, except at most 1, should have non-gaussian distribution and thus it can be viewed as a generalization of PCA to non-gaussian data.
\\ 
\indent It turns out that PCA is a part of ICA. It is used to whiten the data matrix before doing ICA
 which means that, since PCA works with first two moments of distribution, PCA and ICA do not intersect at all. Although, as mentioned by \citeA{aires2000independent}, PCA might provide a good starting point for iterative algorithm of ICA.
The main difference between them is that they are designed to achieve different goals: PCA is designed to maximize variance with certain "bonuses" as orthogonality, linear independence and dimensionality reduction. 
On the other hand, ICA is used for separation of components (\citeA{jolliffe2002principal}), extending this separation to a strict sense of independence than just uncorrelatedness, and it might fail to return reasonable outcome if there is no mixture (linear on non-linear) of independent sources in the data. \\
\indent In contrast to factor analysis, ICA does not try to explain correlations between factors. Rather it assumes that ecologically valid factors will be independent.\\
\indent Both techniques are not robust to noise which is the main weakness when they are applied to financial data. Although several variations exist (see for example \citeA{ikeda2000independent}, \citeA{voss2013fast}, \citeA{bingham2000fast}, \citeA{candes2011robust}) and PCA and ICA might actually be used for noise separation. In particular (based on their properties) ICA is best in non-Gaussian noise separation and PCA in Gaussian respectively.

\section{Comparison}

\begin{paracol}{2}

\begin{leftcolumn}

\section{\underline{PCA}}

\subsection{Inputs}

$X$ - realization of $N$ dimensional random variable. Thus $X$ will be a data matrix $ X \in R^{T \times N}$ with $T$ being the number of samples and $N$ the number of variables (features). $E(X)$ is approximated by sample average

\subsection{Output}

$Z = XW$ - projection of $X$ onto a new basis giving the principal components. Here dimension of $W$ is $N$ by $m : m \leq N$ - by projecting onto $m$ basis vectors the dimensionality of the data might be reduced which is one of the properties of PCA.

\subsection{Problem statement}

For given $X$ find orthogonal set of $m$ basis vectors $w_j$ and corresponding scores $z_i \in R^m $ such that the reconstruction error $V(w,z) = \dfrac{1}{T} \sum_{i=1}^{T} \norm{x_i-z_iW^\intercal}^2 $ is minimized with $W$ being orthonormal, $x_i$ and $z_i$ being row vectors of $X$ and $Z$ respectively and $\Sigma = X^\intercal X$ - sample covariance matrix of $X$. We will see that this is equivalent to maximizing variance of $Z$ that is $J(W,Z)$ (see Appendix).

\subsection{Variance function}

$J(W,Z) = W^\intercal \Sigma W $

\subsection{Assumptions}

\begin{enumerate}
\item $ w_i^\intercal w_j=0 : j \neq i $
\item $\norm{w_i}=1$ for all $i$
\item $ E[X] = 0 $
\end{enumerate}

\subsection{Algorithm}

Variance function above is maximized. Solution will be eigenvectors of $\Sigma$ simply because PCA is equivalent to projecting $X$ onto first $m$ eigenvectors. Refer to Appendix for further details.

\end{leftcolumn}

\begin{rightcolumn}

\section{\underline{ICA}}

\subsection{Inputs}

$X$ - observed $N$ dimensional random variable. On practice $E(X)$ will be approximated by sample average and $X$ will be a data matrix $X \in R^{T \times N}$ with $T$ being the number of realizations of $N$ dimensional discrete-time signal and $N$ the number of variables. 
\\
\noindent $S$ - latent $N$ dimensional (we assume that dimensionality of $X$ and $S$ is the same) random vector which is called a source signal. \\
For notation purposes we assume $X$ and $S$ to be random except for empirical section where we replace it with a realized data matrix.

\subsection{Output}

$Y=W^\intercal X$ - invertible transformation of $X$ giving the estimated source signals.

\subsection{Problem statement}

Blind Separation Problem: given $X$ resulting from linear combination of unknown sources $S$ we need to find a demixing matrix of weights $W$ s.t $Y = W^\intercal X$ is as statistically independent as possible. This is achieved by maximizing the non-gaussianity $\sim$ negentropy $N(Y)$ of Y, which is equivalent (according to Karush–Kuhn–Tucker conditions) to minimizing the objective function $J(W)$ below, where $\varphi(\cdot)$ is the derivative of $\Phi(\cdot)$. The former is a non-quadratic contrast function used to approximate negentropy. For a full set of examples and properties of $\Phi(\cdot)$ see \citeA{hyv1999fast}.

\subsection{Objective function}

$J(W) = E[X \varphi(W^\intercal X)] - \lambda W $

\subsection{Assumptions}

\begin{enumerate}
\item $ w_i^\intercal w_j=0 : j \neq i $
\item $\norm{w_i}=1$ for all $i$
\item $ E[X] = 0 $
\item $X$ is pre-whitened ($X=MX:M^\intercal M = \Sigma^{-1} \rightarrow \Sigma = I$)
\item $S$ are statistically independent (although components of $X$ are dependent since it is a linear combination of $S$ )
\item $W$ is a square matrix (number of observations = number of sources)
\item No noise
\end{enumerate}

\subsection{Algorithm}

Objective function above is minimized. Separation is achieved by projecting $X$ onto the orthonormal basis $W$ such that the resulting estimated sources are fully independent. Usually $W$ are estimated by iteratively minimizing the objective function above or maximizing the non-gaussianity between estimated sources(negentropy). Refer to Appendix for further details.

\end{rightcolumn}

\end{paracol}

\section{Empirical analysis of PCA vs ICA performance}

We based our analysis on underlying assumptions of PCA and ICA and empirically investigated their relative performance by using these algorithms to solve a simple BSS problem, reduce noise, find "interesting" directions in the data and to identify specific situations when one method should be preferred over another. Our main goal was to show the sensitivity of each method to changes in input distribution. \\
In this section we work with realizations of $X$ and $S$ that are now data matrices with dimensionality $T$ by $N$. Thus, as our input sources, we chose two dimensional(for visualization purposes) vector $S = [s_1,s_2]$ : sine function as a first source (except for the last case when it was replaced with a line) $s_{t,1} = sin(\dfrac{t}{10})$ and one-dimensional random noise with gaussian distribution, zero mean, volatility $\sigma$ as a second source such that $s_{t,2}$ is its realization at $t$ for $t=1...T$. Thus we chose $N = 2$ and $T=10000$.

\begin{figure}[H]
\centering
  \includegraphics[scale=0.4]{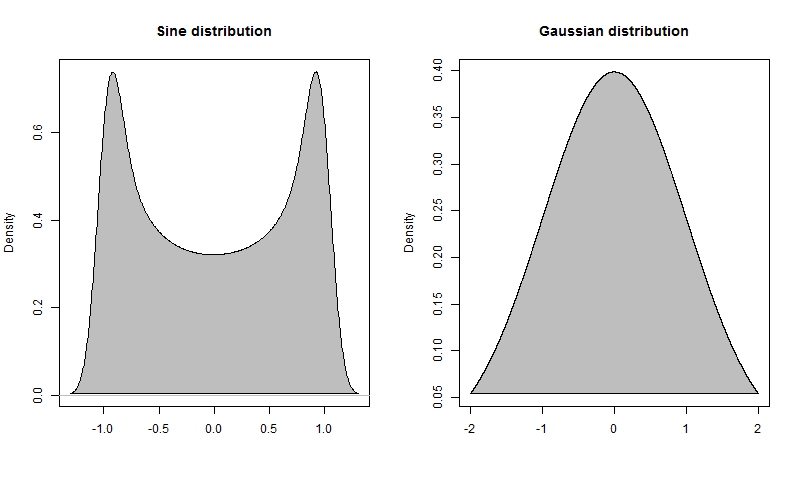}
  \caption{Source distributions}
  \label{dist}
\end{figure}

\noindent To create a mixture of sources $X$ we used the following mixture matrix: 
$$ A = 
\begin{bmatrix}{}
  1.00 & 1.00 \\ 
  -1.00 & 3.00 \\ 
\end{bmatrix}
$$
and thus $$ X = SA $$
\noindent will be a data matrix with $T=10000$ rows and $N=2$ columns. Afterwards, the data was centered before applying PCA and whitened before ICA procedure. We used RMSE to measure the difference between estimated source signals $Y$ and actual sources $S$, both scaled.

\subsection{ICA outperformance}

\begin{figure}[H]
\centering
  \includegraphics[scale=0.55]{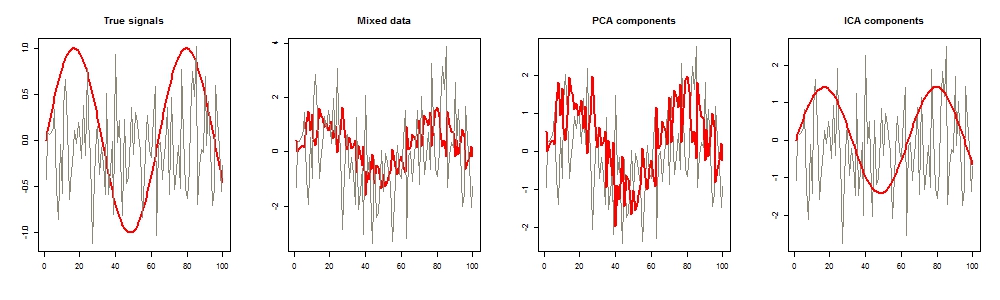}
  \caption{ICA outperformance}
  \label{ica}
\end{figure}

Relying on assumptions of PCA we can conclude that in the absence of pronounced direction in low order moments PCA will fail to partition the mixed signals even if the volatility of one of them is much higher than another. Although this would not be the case for ICA since it relies on higher order moments and will greatly benefit from sine`s negative kurtosis.\\
\indent In our experiment we set $\sigma$ such that the correlation between source signals is zero, plotted the estimated signals $Y$ received from PCA and ICA and compared them with actual $S$. The dominance of ICA is obvious from visual inspection of Figure~\ref{ica}, although we presented the numeric results as well.\\
\indent Generally speaking, since ICA includes PCA as a preprocessing step usually ICA shows better results in terms of signals separation. Hence it is easy to show its relatively higher accuracy :  \citeA{draper2003recognizing} compared PCA and ICA applied to face recognition and found ICA to be superior. Although they failed to explain this difference in terms of input distributions - something we would try to catch up in this paper. Furthermore \citeA{romero2011pca} has studied how well PCA and ICA reduce noise in multi-lead ECG and received better results for the former.\\

\begin{table}[H]
\centering
\begin{tabular}{rrr}
  \hline
 & PCA & ICA \\ 
  \hline
First Source & 0.510 & 0.002 \\ 
  Second Source & 0.521 & 0.013 \\ 
   \hline
\end{tabular}
\caption{ICA outperformance, RMSE} 
\end{table}

\subsection{Similar performance of ICA and PCA}

\begin{figure}[H]
\centering
  \includegraphics[scale=0.55]{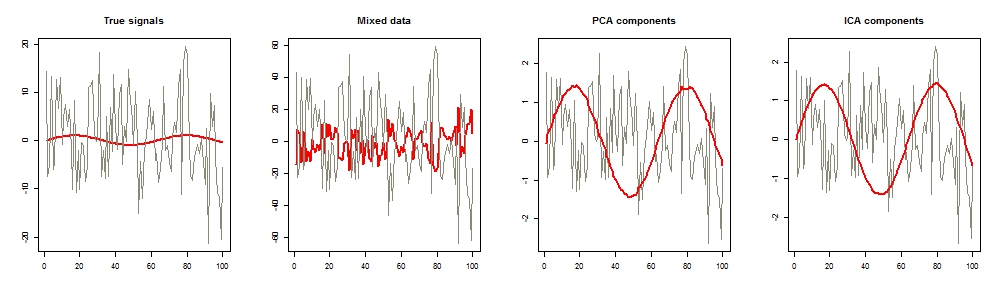}
  \caption{Similar performance}
  \label{similar}
\end{figure}

As soon as directionality (measured by correlation) in source signals increases, PCA gradually improves. In fact, we observe an interesting situation when higher magnitude of correlation means more for better performance than lower noise. On Figure~\ref{ica} one can see that with relatively low level of additive noise but zero correlation PCA fails significantly compared to ICA. However, when on Figure~\ref{similar} we increased the noise level setting $\sigma = 8$ (moving correlation closer to -1) the accuracy of PCA increased a lot.

\begin{table}[H]
\centering
\begin{tabular}{rrr}
  \hline
 & PCA & ICA \\ 
  \hline
First Source & 0.027 & 0.011 \\ 
  Second Source & 0.018 & 0.021 \\ 
   \hline
\end{tabular}
\caption{Similar performance, RMSE} 
\end{table}

\subsection{PCA outperformance}

As discussed earlier, ICA requires that all but one sources should have a non-Gaussian distribution. This is a crucial part since the objective function of the method is measured in terms of non-gaussianity. That being said, we could slightly change our initial setting and think of a simple case when we have one source being a straight line and another one being a random gaussian noise with $\sigma = 10 $. After mixing we will get two gaussian signals.\\
\indent Since PCA takes into account only first and second order moments together with correlations, that should not significantly affect its performance. Standard deviation of a constant is zero and correlation between the mixed signals is one. PCA will simply project all the data on one component. However, ICA will still try to iterate and find maximum separation between two gaussians. That will lead to a slightly worse performance: standard deviation of the estimated source signal that is related to straight line will be $>0$ as well as $RMSE=0.021$. Both higher than those for PCA. This is not a unique case when PCA outperforms ICA. For example \citeA{baek2002pca} showed that algorithm`s competitiveness is significantly determined by the chosen distance metric, concluding that PCA with certain norms might be preferable. \citeA{thomas2002noise} proved that PCA works better than ICA for removal of unstructured random noise.

\begin{figure}[H]
\centering
  \includegraphics[scale=0.55]{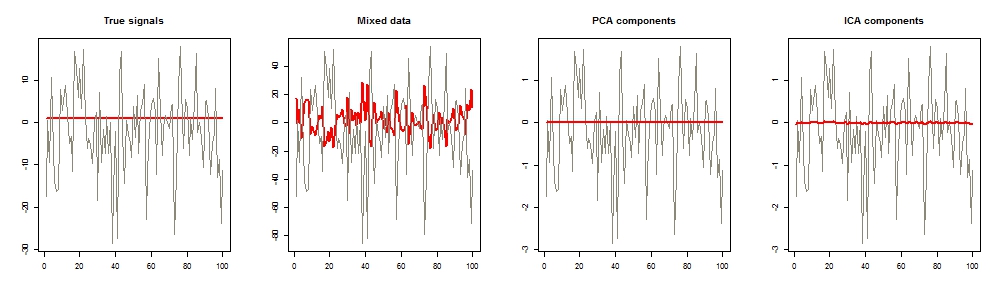}
  \caption{PCA outperformance}
  \label{pca}
\end{figure}

\begin{table}[H]
\centering
\begin{tabular}{rrr}
  \hline
 & PCA & ICA \\ 
  \hline
First Source & 0.000 & 0.021 \\ 
  Second Source & 0.000 & 0.000 \\ 
   \hline
\end{tabular}
\caption{PCA outperformance, RMSE} 
\end{table}

\section{Conclusion}

We performed a comparative study of Principal Component Analysis (PCA) and Independent Component Analysis(ICA) algorithms. Although their natures are quite different, their application to the same problems such as noise reduction, Blind Source Separation, pattern recognition etc. is widespread. The choice of one over another is usually based not only on the kind of problem to be solved, but also on the underlying data distributions. Relying on our empirical analysis and previous research in this area we might sketch a simple rule of thumb: if the data exhibits gaussian distribution, is corrupted by additive gaussian noise and has pronounced directionality (correlation) at place, then PCA might be preferable to ICA. We also found that for PCA high magnitude of correlations plays a more important role than a lower volatility of noise.

\section{Appendix}

\subsection{PCA derivation}

\textbf{Find $w_1$ :}\\

\noindent Following \citeA{murphy2012machine} and taking  $ \dfrac{\partial V(w_1,z_1)}{\partial z_1} $  rewrite the objective function: $$V(w_1)= const - \dfrac{1}{T} \sum_{i=1}^{T} z_{1,i}^2 = const - Var[z_1]$$ hence $$\underset{w_1}{\arg\min} V(w_1) = \underset{w_1}{\arg\max} Var(z_1)$$
Minimizing reconstruction error is similar to maximizing the variance of projected data. By observing that $$\dfrac{1}{T}\sum_{i=1}^{T} z_{1,i}^2 = w_1^\intercal\Sigma w_1 = J(w_1,z_1)$$ and $$\dfrac{\partial \tilde{J}(w_1) }{\partial z_1} = 0 $$
where $\Sigma$ is the covariance matrix of $X$ and $$ \tilde{J}(w_1) = w_1^\intercal \Sigma w_1 + \lambda_1(w_1^\intercal w_1 - 1 )$$
we get $\Sigma w_1 = \lambda_1 w_1$ such that $w_1$ is the eigenvector of $\Sigma$.
\\\\
\textbf{Find $w_j$ : }\\\\
As proven above $w_j$ will be eigenvectors of $\Sigma$.
\\\\
In the literature there is also an alternative representation of PCA technique. $X$ is given as a sum of low-rank matrix $L_0$ and perturbation matrix $N_0$: $$ X = L_0 + N_0 $$ Then the objective function might be formulated as $\norm{X-L}$ and it should be minimized subject to condition $rank(L) \leq k $. For further details see \citeA{candes2011robust}.

\subsection{ICA derivation}

\textbf{Find $w_1$ :}\\

\noindent Because ICA`s goal is to achieve statistical independence in a strict sense, it goes beyond 2nd order moments and demands source signals to be non-gaussian (although one source at most can be gaussian). Thus, following the FastICA procedure in \citeA{hyvarinen2000independent} and \citeA{hyv1999fast} we strive to maximize this non-gaussianity which is measured by negentropy 
\begin{equation*}
\begin{split}
N(X) &= H(X_g) - H(X)\\
 &= E[\Phi(Y)] - E[\Phi(U)]^2
\end{split}
\end{equation*}
\noindent where $X_g$ is a gaussian random variable with the same covariance matrix as that of $X$, $H(X)$ is a differential entropy and $U$ is a random variable with standard normal distribution.
By choosing one of non-quadratic functions $\Phi(Y) = log(cosh(Y))$ and by observing that $U$ is independent of $W$ we can re-write our objective: 
$$\underset{W}{\arg\max} N(Y) \sim \underset{W}{\arg\max} \Phi(W^\intercal X)$$ which is equivalent to the objective function above, where 
$\varphi(Y) = \dfrac{\partial \Phi(Y) }{\partial Y} $.
Finally, using Newton Method and sample average to compute expectations and iterating through $$ w^+ = E[ \varphi'(w^\intercal X)]w - E[X \varphi(w^\intercal X)] $$ where $w^+$ is the new value of w, we can calculate $w_1 = \dfrac{w^+}{\norm{w^+}}$

\noindent\textbf{Find $w_j$ :}\\

\noindent Weights have to be orthonormal which is is achieved by Gram Schmidt orthogonalization. Define a new weight vector as $\alpha_{j}$. Thus on subsequent $T$ samples of $X$ a new weight vector is calculated in the same way as $w$ in one-unit version s.t $j = 2,3...N$. Finally:
$$\theta_{j} = \alpha_{j} - \sum_{i=1}^{j-1} (\alpha_{j}^\intercal w_i)w_i$$ and $$w_{j} = \dfrac{\theta_{j}}{\norm{\theta_{j}}}$$

\bibliography{pca_vs_ica} 
\bibliographystyle{apacite}

\end{document}